\documentclass[11pt]{article}

\usepackage[preprint]{acl}

\usepackage{times}
\usepackage{latexsym}

\usepackage[T1]{fontenc}

\usepackage[utf8]{inputenc}

\usepackage{microtype}

\usepackage{inconsolata}

\usepackage{graphicx}
\usepackage{hyperref}
\usepackage{url}
\usepackage{booktabs}
\usepackage{threeparttable}
\usepackage{tabularx} 
\usepackage{array} 
\usepackage{xurl}
\usepackage{amsmath} 
\usepackage{amssymb}
\usepackage{booktabs}
\usepackage{multirow}
\usepackage[table]{xcolor}

\usepackage[most]{tcolorbox}
\usepackage{xcolor}

\definecolor{linegray}{gray}{0.95}

\tcbset{
  promptstyle/.style={
    colback=linegray,
    colframe=white,
    boxrule=0pt,
    sharp corners,
    left=6pt,
    right=6pt,
    top=4pt,
    bottom=4pt,
    fontupper=\ttfamily\scriptsize,
    before upper=\raggedright,
    breakable,
  }
}
\newtcolorbox{promptbox}[1][]{promptstyle,#1}

\definecolor{IndAcc}{HTML}{D8E2DC}    
\definecolor{DistAlign}{HTML}{E8D5C4} 
\newcommand{\indmetric}[1]{\cellcolor{IndAcc}{#1}}
\newcommand{\distmetric}[1]{\cellcolor{DistAlign}{#1}}
\newcommand{\indcap}[1]{\begingroup\setlength{\fboxsep}{1pt}\colorbox{IndAcc}{#1}\endgroup}
\newcommand{\distcap}[1]{\begingroup\setlength{\fboxsep}{1pt}\colorbox{DistAlign}{#1}\endgroup}

\title{MetaPersona: Task-Grounded Synthetic Populations \\from Empirical Social Science}

\author{
  \textbf{Jinyi Ye\textsuperscript{1}},
  \textbf{Yuangang Li\textsuperscript{2}},
  \textbf{Chenxiao Yu\textsuperscript{1}},
  \textbf{Preyashi Poddar\textsuperscript{1}},
  \textbf{Priyanka Dey\textsuperscript{1}},
  \\
  \textbf{Longtian Ye\textsuperscript{4}},
  \textbf{Zihan Wang\textsuperscript{4}},
  \textbf{Xiyang Hu\textsuperscript{3}},
  \textbf{Emilio Ferrara\textsuperscript{1}},
  \textbf{Yue Zhao\textsuperscript{1}}
\\
  \textsuperscript{1}University of Southern California,
  \textsuperscript{2}University of California, Irvine,
\\
  \textsuperscript{3}Arizona State University,
  \textsuperscript{4}2077AI
\\
  \small{
    \textbf{Correspondence:}
    \href{mailto:jinyiy@usc.edu}{jinyiy@usc.edu},
    \href{mailto:emiliofe@usc.edu}{emiliofe@usc.edu},
    \href{mailto:yue.z@usc.edu}{yue.z@usc.edu}
  }
}

\begin{document}
\maketitle

\begin{abstract} 
Personas used to seed LLM social simulations face a cold-start problem: existing methods lack a principled basis for deciding which attributes to include and how to assign their values. As a result, synthetic populations may misrepresent the demographic composition, latent attributes, and dependency structure that shape downstream behavior. We introduce \textsc{MetaPersona-DB}, a dataset of 11{,}000+ empirical human-subjects studies annotated with task-relevant variables, reported relationships, and aggregate-level population statistics. Building on this resource, we propose \textsc{MetaPersona}, a framework that retrieves task-relevant evidence, constructs literature-derived persona dependency graphs, and samples synthetic populations from empirical priors linking demographics, latent attributes, and outcomes. Across three downstream case studies, three baselines, and three frontier models, results vary by task and model: \textsc{MetaPersona} performs strongly on misinformation belief and AI-tool sentiment, while results on income redistribution are mixed. It also reduces persona-construction cost to under {\$0.5} per task using \texttt{GPT-5.2}. Finally, we present \textsc{MetaPersona-Studio}, a prototype interactive interface for empirically grounded persona generation. \footnote{This manuscript reports work in progress; code and a public demo will be released with the next version.}
\end{abstract}

\section{Introduction}
\label{sec:intro}
Social simulations with large language models (LLM) begins by describing the people to be simulated. Researchers assign a language model a \emph{persona}, a profile of demographic, psychological, and behavioral attributes, and prompt it to respond as that person would: answering a survey item, making a decision, or rendering a judgment \citep{argyle2023one, horton2023homo}. Repeating this process over many personas turns a single model into a simulated population, allowing researchers to approximate aggregate opinions, decisions, and behavioral patterns \citep{park2024generative, yang2024oasis}. This approach has become an active methodological proposal in computational social science, where it promises to reduce the cost and latency of human-subjects research for hypothesis screening, opinion forecasting, and policy analysis \citep{horton2023homo, manning2024automated, sarstedt2024silicon}.

The fidelity of any such simulation is bounded by the population it is built on. If the personas fed to the model misrepresent the target population's composition or internal structure, careful prompting at simulation time cannot recover it \citep{li2026promise}. Persona generation at scale is therefore a central bottleneck. Existing work samples structured profiles from census marginals \citep{ge2024scaling}, prompts LLMs to elaborate free-form biographies \citep{moon2024anthology}, or mines narrative personas from social media \citep{hu2025population}. Yet the field still lacks what \citet{li2026promise} call a \emph{science of persona generation}. Current methods remain heuristic, leaving two core questions unresolved: which attributes should a task-specific persona include, and how should those attributes be structured to reflect real population distributions and empirical dependencies?

\begin{figure*}[t]
\centering
  \includegraphics[width=\linewidth]{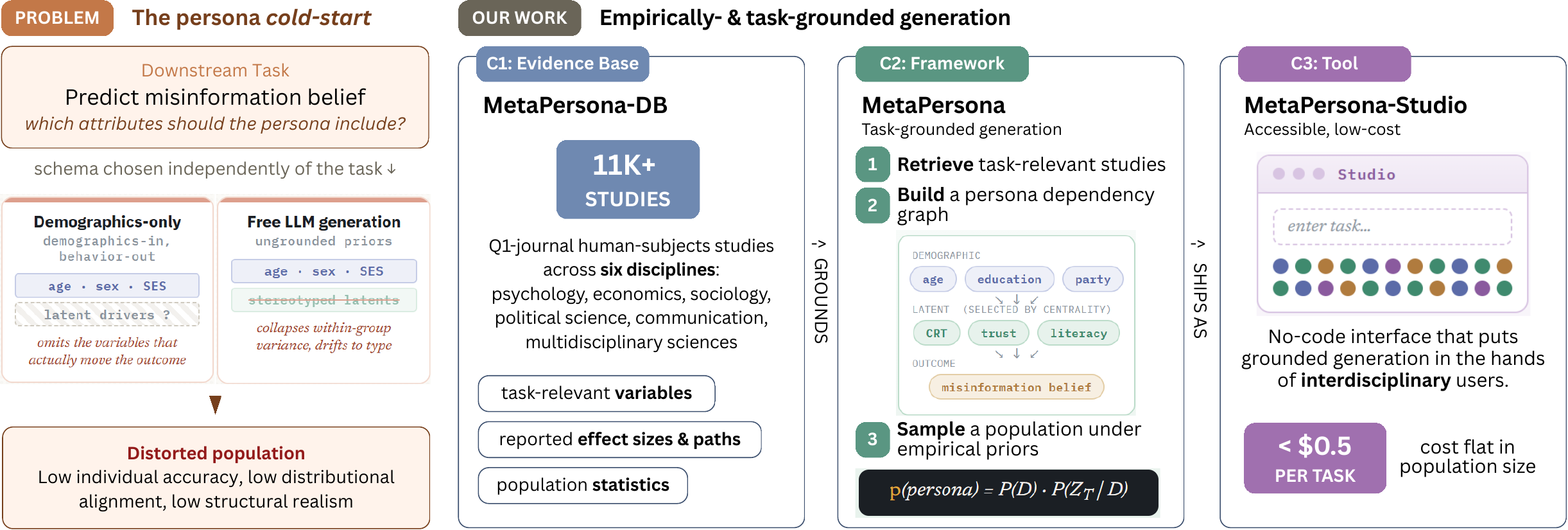}
  \caption{%
    \textbf{The persona cold-start problem and our response.}
    We address the problem through three contributions:
    \textsc{MetaPersona-DB}, a curated empirical evidence base for persona variables;
    \textsc{MetaPersona}, a framework for selecting task-relevant attributes and sampling empirically grounded synthetic populations;
    and \textsc{MetaPersona-Studio}, an accessible low-cost interface for interdisciplinary use.%
  }
  \label{fig:overview}
  \vspace{-0.1in}
\end{figure*}

\textbf{Gap 1: The persona cold-start problem.} Before simulation begins, researchers must decide which attributes a persona should contain, but this choice is often made by default rather than design. We call this the \emph{persona cold-start problem}. Existing pipelines improve distributional alignment \citep{hu2025population}, topic-adaptive sampling \citep{chen2026hag}, and narrative richness \citep{moon2024anthology}, but typically start from a fixed or LLM-nominated schema chosen independently of the downstream task. Most default to what \citet{li2026simulating} call the \emph{demographics-in, behavior-out} paradigm, treating a handful of demographic fields as proxies for human behavior. Yet empirical social science shows that the attributes that matter depend heavily on the outcome: ideology and partisanship shape vote choice \citep{argyle2023one, yu2025large}, while cognitive reflection predicts misinformation susceptibility \citep{pennycook2019lazy}, and demographics alone may explain only a small fraction of variance in such outcomes \citep{hu2024quantifying}. 

\textbf{Contribution 1: A dataset and framework for empirically- and task-grounded persona generation.} We address the persona cold-start problem by grounding attribute selection in empirical literature rather than heuristics. We curate \textsc{MetaPersona-DB}, a dataset of 11{,}000+ Q1-journal human-subjects studies across six social science disciplines, produced through a nearly \$2{,}000 annotation process that converts scattered empirical evidence into structured variables, effect sizes, and population statistics for persona generation. Building on this resource, we introduce \textsc{MetaPersona}, a framework that retrieves task-relevant studies, organizes extracted variables into a persona dependency graph linking demographic and latent factors to the target outcome, and samples synthetic populations using empirically derived priors. The dataset and framework specify \emph{which} attributes a task requires and \emph{how} to assign values that reflect real population structure.

\vspace{-0.05cm}
\textbf{Gap 2: Realism of persona-driven downstream tasks.} Identifying the right attributes is necessary but not sufficient; the test is whether they improve downstream behavioral fidelity. Naive LLM-generated personas can distort downstream outcomes by collapsing within-group variability \citep{wang2025flatten}, exaggerating associations between variables \citep{xie2026evaluating}, and drifting toward stereotyped or homogeneous responses \citep{bisbee2024perils, li2026promise}. Whether our persona construction mitigates these failures remains an open empirical question.

\vspace{-0.05cm}
\textbf{Contribution 2: Cross-model evaluation of downstream realism.} We evaluate \textsc{MetaPersona} on three cross-disciplinary case studies against three baselines that mirror current practice: demographics-only sampling, demographics plus LLM-inferred latent attributes, and life-backstory generation \citep{moon2024anthology}. We run this evaluation across three frontier models: \texttt{GPT-5.2}, \texttt{Claude~Haiku~4.5}, and \texttt{DeepSeek-V3.2}. Following recent practice, we measure three complementary dimensions of realism: individual-response accuracy, population-level distributional alignment, and structural realism, defined as the preservation of associations and predictive structure needed for social inference \citep{xie2026evaluating, suh2025language}. Results vary across tasks and models: gains are strongest for misinformation belief and AI-tool sentiment, whereas the income-redistribution results are mixed and include settings in which baselines perform better.

\vspace{-0.05cm}
\textbf{Gap 3: Cost and accessibility for interdisciplinary use.} Persona-based simulation is relevant not only to NLP but also to the social science communities that would deploy it \citep{li2026promise}. Yet adoption is limited by cost: generating large synthetic populations by prompting frontier models can be expensive and technically demanding.

\vspace{-0.05cm}
\textbf{Contribution 3: An accessible and low-cost persona generation tool.} Because \textsc{MetaPersona} is predominantly statistical, sampling from an empirically grounded graph and requiring only minimal model calls, it generates a full population for under \$0.5 per task, 10--50$\times$ below other LLM-based baselines. After the graph is constructed, sampling cost is essentially independent of population size (see Appendix~\ref{appendix:cost-analysis}). We present \textsc{MetaPersona-Studio}, a prototype interactive interface designed for interdisciplinary users. Figure~\ref{fig:overview} summarizes the research gaps and our contributions.

\section{Related Work}
\label{sec:related-work}
\vspace{-0.1in}

\textbf{Naive persona generation distorts simulations.}
LLM personas built from shallow or ad hoc attributes can misrepresent the populations they stand for. Prior work shows that scalable persona generation introduces systematic biases \citep{li2026promise}, flattens identity groups \citep{wang2025flatten}, and regresses toward typical profiles regardless of scale \citep{xie2026evaluating}. These failures reflect a broader ``demographics-in, behavior-out'' paradigm \citep{li2026simulating}, which jumps from coarse demographics to predicted outcomes.

\textbf{Existing responses lack empirically grounded latent structure.}
Most realism-oriented methods keep the schema fixed and improve the assigned \emph{values}, for example by aligning attributes to reference distributions \citep{hu2025population} or enriching personas with free-text backstories \citep{moon2024anthology, park2024generative}. Yet \citet{hu2024quantifying} show that persona prompting helps only when the chosen attributes predict the outcome, and existing persona variables explain under 10\% of annotation variance. Better values cannot compensate for an uninformative schema, yet these approaches do not determine which latent attributes a task requires.
Concurrent work introduces task-sensitivity into persona generation. \citet{chen2026hag} builds a topic-adaptive hierarchy over demographic attributes using a world knowledge model, while \citet{li2026promise} sample a demographic base from census data and expand it through LLM prompting. Both show the value of structured, task-sensitive persona generation.

\textbf{From demographic base to task-grounded latents.}
\textsc{MetaPersona} extends this direction by adding latent psychological attributes between demographics and outcomes. Rather than relying only on LLM-derived priors, we retrieve task-relevant empirical studies and use their reported relationships to derive the latent layer: selecting task-relevant attributes and sampling their values conditioned on demographics under literature-derived priors. This grounds the schema itself in measured social science evidence before values are assigned. A fuller discussion appears in Appendix~\ref{appendix:extended-related-work}.

\section{\textsc{MetaPersona-DB}: A Variable-level Empirical Social Science Dataset}
\label{sec:dataset}
\vspace{-0.1in}

A persona is only as good as the attributes it encodes, and which attributes matter is an empirical question that social science has spent decades answering. The first step toward a science of persona generation is therefore to mine this evidence systematically. We do so by curating \textsc{MetaPersona-DB}, a variable-level dataset distilled from thousands of empirical human-subjects studies.

\begin{figure}[t]
  \centering
  \includegraphics[width=0.85\linewidth]{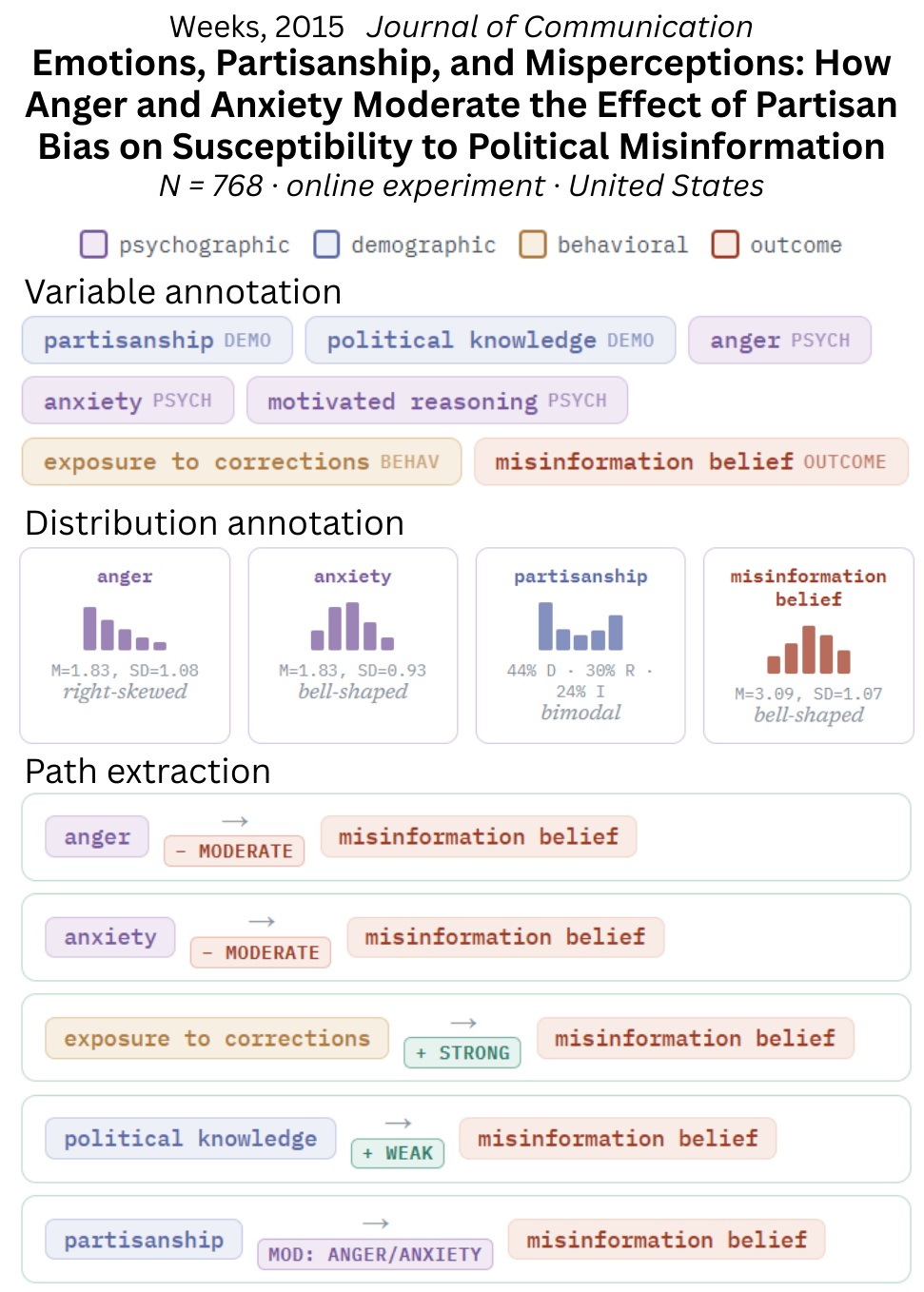}
  \caption{An example of an annotated paper.}
  \vspace{-0.2in}
  \label{fig:paper_example}
\end{figure}

\vspace{-0.1in}
\subsection{Paper Selection}

\textsc{MetaPersona-DB} is built from empirical studies across six social science disciplines indexed by Journal Citation Reports (JCR; \citet{clarivate2024jcr}): psychology, political science, economics, sociology, communication, and multidisciplinary sciences. We select these fields because they represent core domains in which LLM-based social simulations are increasingly applied \cite{gao2024large, ziems2024can}, a primary use case for \textsc{MetaPersona}. We restrict the corpus to publications after 2000 in first-quartile (Q1) journals indexed by JCR, with at least 20 citations. This filtering strategy prioritizes studies with recognized journal quality, methodological credibility, and demonstrated scholarly influence \cite{hirsch2005index}. Paper metadata are retrieved through the OpenAlex API \cite{priem2022openalex}. 
 
\subsection{Identifying Empirical Human Studies}
\vspace{-0.05in}
Not all papers in social science journals report original human data: many are theoretical, methodological, or narrative reviews that yield no extractable behavioral variables or causal paths. We therefore restrict the corpus to \textit{empirical human-involving studies}---original studies using real-world human data (surveys, experiments, administrative records, etc.) or systematic reviews and meta-analyses synthesizing such studies \cite{reichardt2002experimental, borenstein2021introduction}. Classification was performed using \texttt{gpt-5.2} with a structured prompt, validated by two independent human annotators (see Appendix \ref{appendix:prompt} and \ref{appendix:human-validation}). This yields \textbf{11,638 empirical studies} forming the basis of \textsc{MetaPersona-DB}.

\vspace{-0.05in}
\subsection{Variable and Path Extraction}
\vspace{-0.05in}
From each paper, we extract persona-relevant information from the title and abstract in three steps, using \texttt{gpt-5.2}. A detailed annotation taxonomy is presented in Appendix \ref{appendix:dataset}.

\textbf{Step 1: Variable annotation.} We identify each individual-level \textit{variable}---any attribute that can be measured for a single person. We classify each variable as demographic, psychographic, or behavioral, following prior practices \citep{pruitt2003personas}. For each variable, we annotate how it is defined and measured (e.g., scale score, behavioral count, categorical label).

\textbf{Step 2: Distribution annotation.} We assign each variable a coarse population shape, like bell-shaped, right-skewed, or bimodal distributions. Abstracts rarely report exact means, variances, or full distributions. We therefore treat these shapes as construct-level priors elicited from the annotation model, rather than as study-specific statistics \cite{zhu2025eliciting, capstick2025autoelicit}. When ground-truth population distributions are available, we use them to validate the annotated shapes.

\textbf{Step 3: Path extraction.} We extract directed relationships between variables and encode them as \textit{paths}. Each path links a source variable to a target variable (e.g., risk perception $\rightarrow$ vaccination intention) and records the relationship direction (positive or negative) and effect size (e.g., correlation $r$, regression coefficient $\beta$, odds ratio, mean difference). We map effect sizes to coarse bands (weak, moderate, or strong) using conventional guidelines \cite{cohen2013statistical}. If an abstract reports only the direction, as most do, we keep that direction and use a model-elicited prior for strength. Human evaluators audit a sample of full-text PDFs to check whether these priors are supported by the paper. Repeated paths are then merged into a consensus estimate that accounts for between-study heterogeneity \cite{borenstein2009effect}. Figure \ref{fig:paper_example} shows an example of an annotated paper \citep{weeks2015emotions}.

\textbf{Annotation prompts and human validation.}
All LLM annotation prompts are provided in Appendix~\ref{appendix:prompt}. Human validation is integrated into the steps above: evaluators audit a sample of full-text PDFs to check whether the model-extracted variables, distributions, paths, and priors are supported by the papers. We describe the validation protocol in Appendix~\ref{appendix:human-validation}.

\section{\textsc{MetaPersona}: Task-Grounded Synthetic Population Generation}
\label{sec:metapersona}

\begin{figure*}[t]
\centering
  \includegraphics[width=\linewidth]{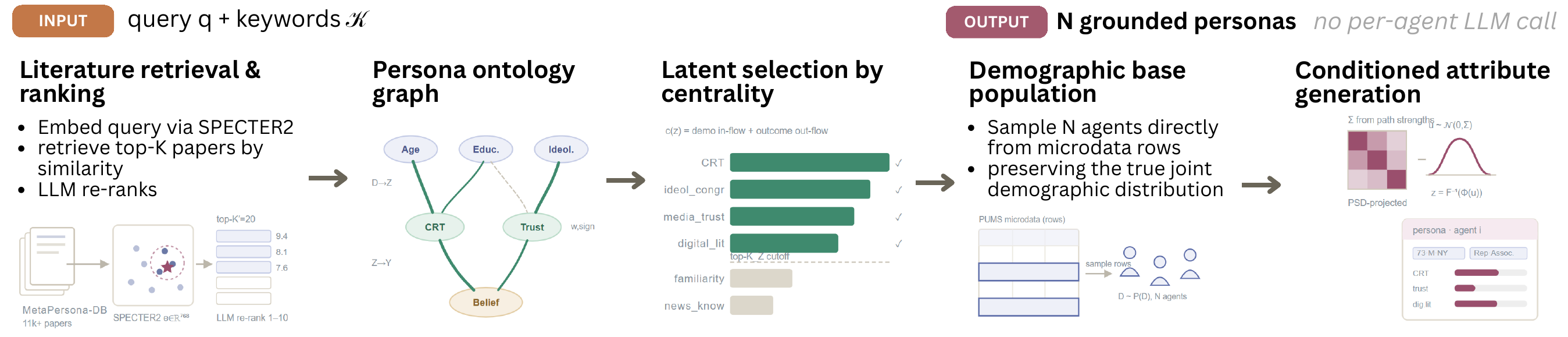}
  \caption{Overview of the \textsc{MetaPersona} pipeline: from a task query to a synthetic population.}
  \label{fig:metapersona-pipeline}
  \vspace{-0.1in}
\end{figure*}

Given a downstream task, \textsc{MetaPersona} generates a synthetic population in five stages. It retrieves task-relevant studies from \textsc{MetaPersona-DB}, builds a task-specific persona graph from extracted variable paths, selects the latent attributes most central to the task outcome, samples a demographic base population, and generates each agent's latent attributes conditioned on its demographics under empirically grounded priors. We describe each stage below, with Figure \ref{fig:metapersona-pipeline} illustrating the key steps.

\vspace{-0.05in}
\paragraph{Preliminary: Task Representation and Embedding.} We embed each paper in \textsc{MetaPersona-DB} using its title and abstract with SPECTER2, a scientific document embedding model designed for scholarly retrieval~\cite{cohan2020specter, singh2023scirepeval}. We use \texttt{allenai/specter2\_base} with the \texttt{allenai/specter2} adapter and take the CLS-token representation $\mathbf{e}_i \in \mathbb{R}^{768}$ as the paper-level embedding. These embeddings are precomputed and cached for the full corpus.

\vspace{-0.05in}
\subsection{Literature Retrieval and Ranking}
\vspace{-0.05in}

A task is specified by a natural-language query $q$ and a small keyword set
$\mathcal{K}$, provided by the researcher.
We encode $q$ with the same SPECTER2 encoder and score each paper using
semantic similarity and keyword overlap:
\begin{equation}
  s_i \;=\; \cos\!\bigl(\mathbf{e}_q,\,\mathbf{e}_i\bigr)
           \;+\; \lambda\, m_i(\mathcal{K}),
  \label{eq:hybrid-score}
\end{equation}
where $m_i(\mathcal{K})$ counts keyword occurrences in paper~$i$ and
$\lambda$ is a small weight ($\lambda{=}0.05$ in our experiments).
We retain the top-$K$ papers by $s_i$, then use an LLM to re-rank them
by task relevance on a $1$--$10$ scale~\citep{sun2023chatgpt}.
The top-$K'$ papers form the retrieval set $\mathcal{R}$
($K{=}100$, $K'{=}20$).
This two-stage design~\citep{nogueira2019passage} uses embeddings for
broad recall and the LLM only for precision ranking over a small
candidate pool.

\vspace{-0.05in}
\subsection{Persona Ontology Graph Construction}
\vspace{-0.05in}

From the retrieval set $\mathcal{R}$, we collect all extracted variable paths and merge them into a task-specific dependency graph $G=(V,E)$. An LLM clusters synonymous variables across papers into a single shared node (e.g., \emph{analytical thinking}, \emph{need for cognition}, and \emph{CRT} collapse into one latent node). 
Each node is assigned a tier $\tau(v)\in\{\textsf{demographic},\,\textsf{latent},\,\textsf{outcome}\}$. A directed edge $u\!\to\!v$ aggregates all paths between two themes. Its weight counts supporting paths, and its sign is the majority-vote direction:
\begin{equation}
\begin{aligned}
  w_{uv} &= \sum_{p \in \mathcal{R}} \mathbb{1}[\,p: u \to v\,], \\
  d_{uv} &= \operatorname*{maj}_{p:\,u\to v}\;\mathrm{dir}(p)
  \in \{+,-\}.
\end{aligned}
\label{eq:edge-weight}
\end{equation}
We retain two edge types for generation: $\textsf{demographic}\to\textsf{latent}$, which captures how demographics shape latent attributes, and $\textsf{latent}\to\textsf{outcome}$, which captures how latents relate to the task outcome. This yields the layered structure $D \to Z \to Y$.

\vspace{-0.05in}
\subsection{Latent Attribute Selection}
\vspace{-0.05in}

The graph often contains more latent themes than a persona needs. We rank each latent node $z$ by a centrality score that favors attributes linked to both demographics and the task outcome:
\begin{equation}
  c(z) \;=\;
    \underbrace{
      \sum_{\substack{u:\,\tau(u)=\textsf{demo}}} w_{uz}
    }_{\text{demographic in-flow}}
  \;+\;
    \underbrace{
      \sum_{\substack{y:\,\tau(y)=\textsf{outcome}}} w_{zy}
    }_{\text{outcome out-flow}}.
  \label{eq:centrality}
\end{equation}
We select the top-$K_Z$ latents by $c(z)$ as the persona's latent schema $Z_T$ (e.g., $K_Z{=}3$). 

\subsection{Demographic Base Population Sampling}
\label{subsec:demo-sampling}

We instantiate a base population of $N$ agents by sampling demographics
$D$ from a researcher-specified grounding source, such as U.S.\ adults
or eligible voters.
For benchmarked tasks, we sample from the ground-truth survey population. Otherwise, we draw from public census microdata, using the American Community Survey Public Use Microdata Sample (PUMS) for U.S.\ populations and IPUMS-International for cross-national targets
\citep{acs_pums, ipums_international}.
Because these sources provide individual-level records rather than marginal summary tables, sampling rows directly preserves the empirical demographic \emph{joint} distribution by construction, avoiding incongruous attribute combinations that arise when joints are reconstructed from marginals \cite{li2026promise}.
Researchers can customize the population through standard filters such as country, age, education, and employment in \textsc{MetaPersona-Studio} (\S~\ref{sec:metapersona-studio}).

\begin{table*}[t]
\centering
\small
\caption{Downstream evaluation tasks, ground-truth data sources, and predicted outcomes.}
\label{tab:tasks}
\begin{tabularx}{\linewidth}{
  >{\raggedright\arraybackslash}p{0.17\linewidth}
  >{\raggedright\arraybackslash}p{0.17\linewidth}
  >{\raggedright\arraybackslash}p{0.19\linewidth}
  >{\raggedright\arraybackslash}X
}
\toprule
\textbf{Task} & \textbf{Discipline} & \textbf{Dataset} & \textbf{Predicted Outcome Variable} \\
\midrule
Misinformation belief
  & Communication, political science
  & MIST \citep{maertens2024misinformation}
  & Ability to distinguish real from fake headlines, measured by numerical MIST-20 score \\
Income redistribution attitude
  & Economics, sociology
  & ISSP 2019 Social Inequality V (ZA7600, v3.0.0)
  & Agreement that government should reduce income differences, measured by ordinal scale \\
AI-tool sentiment \& adoption
  & Behavioral science, HCI
  & Stack Overflow Survey \citep{stackoverflow_survey}
  & Sentiment toward AI tools and expected future use, measured by ordinal scales \\
\bottomrule
\end{tabularx}
\end{table*}

\begin{figure*}[t]
  \includegraphics[width=\linewidth]{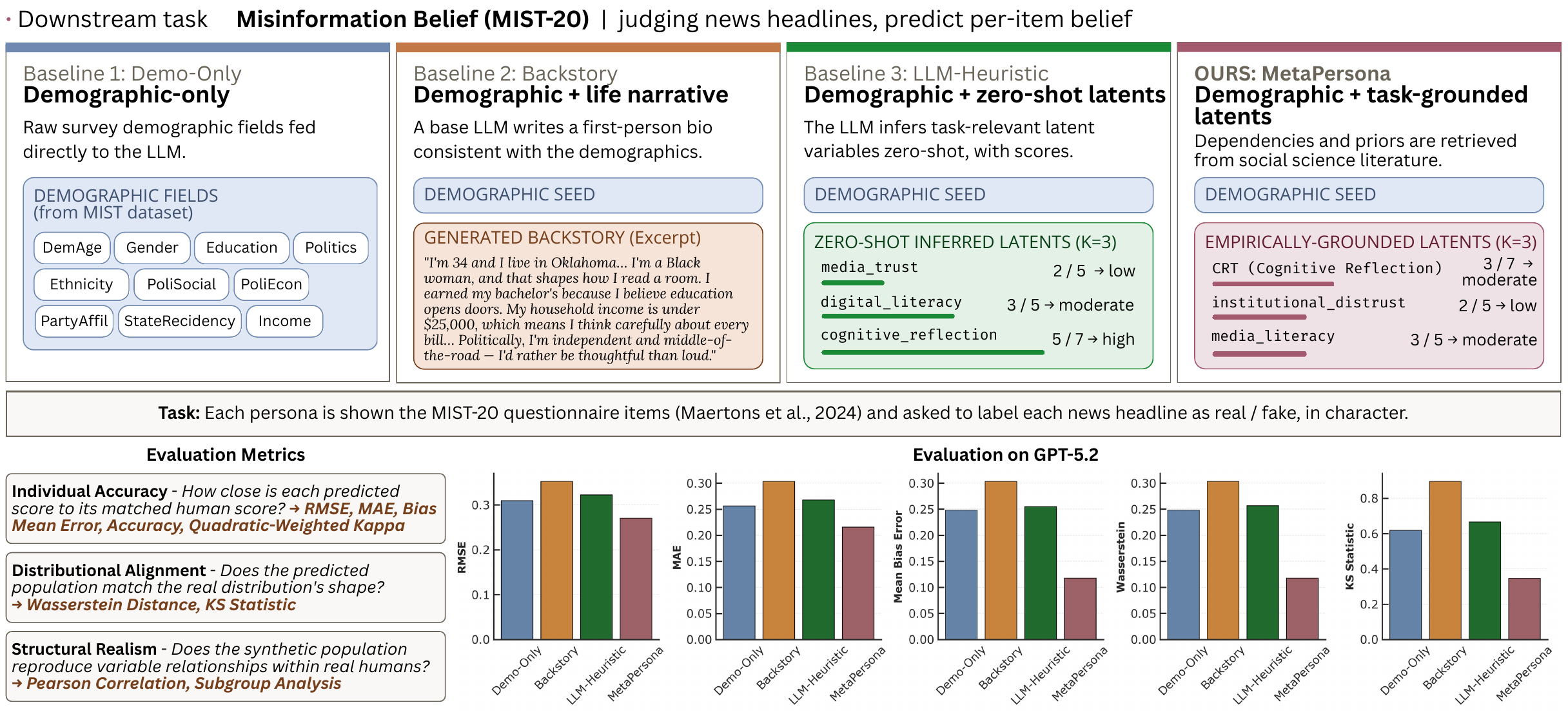}
  \caption{Examples of the four persona-construction methods evaluated for misinformation-belief prediction: \textsc{Demo-Only}, \textsc{Backstory}, \textsc{LLM-Heuristic}, and \textsc{MetaPersona}. Evaluation uses matched responses from real MIST survey participants ($N=1{,}000$) from \citet{maertens2024misinformation}.}
  \label{fig:case-study-demo}
  \vspace{-0.1in}
\end{figure*}

\subsection{Population-Conditioned Attribute Generation}
\label{subsec:generation}

Each agent's persona factorizes into demographics and task-relevant
latents conditioned on those demographics:
\begin{equation}
  p(\text{persona}) \;=\; P(D)\,\cdot\, P(Z_T \mid D).
  \label{eq:persona-factorization}
\end{equation}
To estimate $P(Z_T \mid D)$, we use the dependency structure from the
persona graph.
We sample the joint latent vector with a Gaussian copula, which separates
each variable's marginal distribution from its dependence with other
variables:
\begin{equation}
  \mathbf{u} \;\sim\; \mathcal{N}(\mathbf{0},\,\Sigma),
  \qquad
  z_j \;=\; F_j^{-1}\!\bigl(\Phi(u_j)\bigr),
  \label{eq:copula}
\end{equation}
where $\Phi$ is the standard normal CDF, $F_j^{-1}$ is the quantile
function for latent $j$, and $\Sigma$ is the correlation matrix induced
by reported and elicited path strengths.
Conditioning on each agent's demographics then yields its latent values.
Because generation is mostly statistical, \textsc{MetaPersona} does not
require per-agent LLM calls, making it substantially cheaper than
baselines that generate each persona through prompting.

\section{Evaluation of \textsc{MetaPersona} on Downstream Tasks: Three Case Studies}
\label{sec:case_studies}

\begin{table*}[t]
\centering
\caption{%
  Performance of all methods across three case studies and three LLM backbones.
  Within each base LLM, we compare four persona-construction methods.
  Best result per task and metric in \textbf{bold}; second-best is \underline{underlined}.
  $\downarrow$ = lower is better; $\uparrow$ = higher is better.
  \indcap{Individual accuracy} metrics and \distcap{distributional alignment} metrics are highlighted in the header.
  Bias is signed mean error; best/second-best for Bias are based on smallest absolute bias.
  MIST metrics use scores normalized to $[0,1]$; ISSP MAE and W-dist use the original 1--5 ordinal codes.
  Metric magnitudes should therefore not be compared across tasks.
  For Case Study 3, W-dist is macro-averaged over two predicted outcomes.
}

\scriptsize
\setlength{\tabcolsep}{2pt}
\begin{tabular}{ll ccccc cccc cccc}
\toprule
& & \multicolumn{5}{c}{\textbf{Case Study 1: Misinformation (MIST)}}
  & \multicolumn{4}{c}{\textbf{Case Study 2: Redistribution (ISSP)}}
  & \multicolumn{4}{c}{\textbf{Case Study 3: AI Tool Sentiment (SO)}} \\
\cmidrule(lr){3-7}\cmidrule(lr){8-11}\cmidrule(lr){12-15}
\textbf{Backbone} & \textbf{Method}
  & \indmetric{RMSE$\downarrow$} & \indmetric{MAE$\downarrow$} & \distmetric{Bias} & \distmetric{W-dist$\downarrow$} & \distmetric{KS$\downarrow$}
  & \indmetric{MAE$\downarrow$} & \indmetric{W-1 Acc$\uparrow$} & \indmetric{QWK$\uparrow$} & \distmetric{W-dist$\downarrow$}
  & \indmetric{MAE$\downarrow$} & \indmetric{W-1 Acc$\uparrow$} & \indmetric{QWK$\uparrow$} & \distmetric{W-dist$\downarrow$} \\
\midrule

\multirow{4}{*}{GPT-5.2}
  & Demo-Only            & \underline{0.310} & \underline{0.257} & \underline{0.248} & \underline{0.248} & \underline{0.619}
                         & \textbf{0.812} & \underline{0.817} & 0.086 & \underline{0.583}
                         & \underline{0.248} & \underline{0.743} & 0.059 & \underline{0.808} \\
  & Backstory            & 0.352 & 0.304 & 0.303 & 0.303 & 0.896
                         & 0.828 & \textbf{0.819} & \underline{0.111} & 0.627
                         & 0.250 & \textbf{0.749} & 0.036 & 0.863 \\
  & LLM-Heuristic        & 0.322 & 0.268 & 0.255 & 0.257 & 0.666
                         & 0.842 & 0.808 & \textbf{0.128} & \textbf{0.570}
                         & 0.252 & 0.735 & \underline{0.073} & 0.871 \\
  & \textsc{MetaPersona} & \textbf{0.270} & \textbf{0.216} & \textbf{0.118} & \textbf{0.118} & \textbf{0.346}
                         & 0.832 & 0.814 & 0.087 & 0.642
                         & \textbf{0.217} & \underline{0.743} & \textbf{0.320} & \textbf{0.616} \\
\addlinespace

\multirow{4}{*}{\shortstack{Claude-\\Haiku 4.5}}
  & Demo-Only            & \underline{0.238} & \underline{0.192} & 0.151 & 0.153 & \underline{0.475}
                         & 0.884 & 0.802 & \textbf{0.170} & \underline{0.416}
                         & 0.293 & 0.667 & 0.052 & 0.931 \\
  & Backstory            & 0.254 & 0.203 & 0.171 & 0.172 & 0.527
                         & \underline{0.881} & \underline{0.809} & \underline{0.157} & 0.444
                         & 0.276 & \underline{0.696} & 0.063 & 0.900 \\
  & LLM-Heuristic        & 0.242 & 0.194 & \underline{0.149} & \underline{0.151} & 0.493
                         & 0.978 & 0.749 & 0.116 & \textbf{0.360}
                         & \underline{0.271} & 0.674 & \underline{0.066} & \textbf{0.736} \\
  & \textsc{MetaPersona} & \textbf{0.212} & \textbf{0.171} & \textbf{0.034} & \textbf{0.065} & \textbf{0.198}
                         & \textbf{0.863} & \textbf{0.810} & 0.110 & 0.524
                         & \textbf{0.230} & \textbf{0.725} & \textbf{0.308} & \underline{0.763} \\
\addlinespace

\multirow{4}{*}{\shortstack{DeepSeek-\\V3.2}}
  & Demo-Only            & 0.288 & 0.235 & 0.225 & 0.225 & 0.615
                         & 1.146 & 0.663 & 0.134 & 0.630
                         & 0.288 & 0.677 & 0.085 & \underline{0.477} \\
  & Backstory            & \underline{0.271} & \underline{0.217} & \underline{0.199} & \underline{0.199} & \underline{0.554}
                         & \underline{0.988} & \underline{0.740} & \textbf{0.171} & \underline{0.371}
                         & \underline{0.263} & \underline{0.712} & \underline{0.089} & 0.773 \\
  & LLM-Heuristic        & 0.317 & 0.265 & 0.257 & 0.257 & 0.655
                         & 1.130 & 0.673 & 0.134 & 0.774
                         & 0.287 & 0.614 & 0.074 & 0.605 \\
  & \textsc{MetaPersona} & \textbf{0.242} & \textbf{0.192} & \textbf{0.045} & \textbf{0.045} & \textbf{0.150}
                         & \textbf{0.972} & \textbf{0.744} & \underline{0.144} & \textbf{0.364}
                         & \textbf{0.209} & \textbf{0.743} & \textbf{0.402} & \textbf{0.466} \\

\bottomrule
\end{tabular}
\label{tab:main-results}
\end{table*}

Generating an empirically grounded population is only useful if the resulting personas behave realistically on downstream tasks of genuine social-scientific interest.
We therefore ask: \textit{how well do \textsc{MetaPersona} populations reproduce real human responses, and how do they compare to existing persona-construction practices?} 
We answer this through three case studies, each evaluated against an individual-level human ground truth and against three frontier LLMs (\texttt{GPT-5.2}, \texttt{Claude~Haiku~4.5}, and \texttt{DeepSeek~V3.2}) to test the robustness of our findings.

\vspace{-0.05in}
\subsection{Task Selection and Data Sources}
\vspace{-0.05in}

We evaluate \textsc{MetaPersona} on three downstream tasks chosen against two requirements: each task must be (1) \emph{well studied in the empirical social sciences}, so that \textsc{MetaPersona-DB} contains relevant attribute--outcome evidence, and (2) backed by a \emph{real individual-level survey} recording both demographics and outcomes, so that each synthetic persona can be matched to a real respondent and scored against an observed response. The three tasks span distinct disciplines and outcome types to test generalization across literatures and response formats. Because each persona is matched to a real respondent, demographics are drawn directly from each ground-truth dataset rather than from PUMS/IPUMS (\textit{cf.} \S~\ref{subsec:demo-sampling}). Table~\ref{tab:tasks} summarizes the tasks, data sources, and predicted outcomes; the full list of demographic inputs and inferred latent attributes is in Appendix~\ref{appendix:extended-results}.

\vspace{-0.05in}
\subsection{Baseline Construction}
\vspace{-0.05in}

Because task-grounded, population-level persona generation is, to our knowledge, previously understudied; we therefore construct three
that share the same demographic base population, with an example demonstrated in Figure \ref{fig:case-study-demo}.
(1)~\textbf{Demographics-Only (\textsc{Demo-Only})} conditions each agent on demographics alone and queries the outcome directly, echoing the ``demographics-in, behavior-out'' paradigm \citep{li2026simulating} adopted by most LLM social simulators \cite{gao2023s3, yang2024oasis}.
(2)~\textbf{Demographics + Life Narrative (\textsc{Backstory})} adapts
\textsc{Anthology}~\citep{moon2024anthology}, which generates free-form first-person backstories from demographic profiles, demonstrating that narrative richness beats flat demographic schemas.
(3)~\textbf{Demographics + Zero-Shot Latents (\textsc{LLM-Heuristic})} prompts the LLM to identify task-relevant latent variables and infer their values from demographics alone, without grounding. Inspired by \citet{chen2026hag}, this baseline extends topic-relevant demographic priors to latent variables by letting the LLM nominate and assign task-specific attributes. In contrast, \textsc{MetaPersona} selects latent attributes from \textsc{MetaPersona-DB} and assigns values conditioned on demographics under literature-derived priors.
For a fair comparison, \textsc{LLM-Heuristic} and \textsc{MetaPersona} infer the same number of latent attributes ($K{=}3$).

\subsection{Realism Metrics}

We evaluate realism at three levels---individual accuracy, distributional alignment, and structural realism---so that a method cannot appear faithful merely by matching one aspect of the data while distorting another. 

\textbf{Individual accuracy.}
We measure how closely each synthetic persona reproduces its matched respondent's outcome. For the continuous MIST score, we report RMSE and MAE. For ordinal outcomes, we report MAE, within-one accuracy (W-1 Acc), and quadratic-weighted kappa (QWK).

\textbf{Distributional alignment.}
We measure whether the predicted population reproduces the shape of the human response distribution. We report Wasserstein distance for all tasks and the mean signed bias and Kolmogorov--Smirnov statistic (KS) for the continuous MIST outcome. For Stack Overflow, Wasserstein distance is macro-averaged over two predicted outcomes, \texttt{AISent} and \texttt{AISelect}.

\textbf{Structural realism.}
We assess whether synthetic populations reproduce relationships observed in human data. We examine whether predicted outcomes recover literature-supported associations.
Detailed metric definitions and calculation procedures are provided in Appendix~\ref{appendix:experiment-design}.

\subsection{Aggregated Results and Cross-Task Analysis}

\textbf{Individual accuracy.}
As shown in Table~\ref{tab:main-results}, results vary by task and backbone. \textsc{MetaPersona} achieves the best RMSE/MAE on MIST under all three models and the lowest ISSP MAE with \texttt{Claude~Haiku~4.5} and \texttt{DeepSeek-V3.2}, but not with \texttt{GPT-5.2}, where \textsc{Demo-Only} has the lowest MAE. On ISSP, different baselines also lead on W-1 Acc and QWK depending on the backbone, making this a mixed rather than uniformly positive result. On SO, QWK rises from the best baseline value of $0.073$ to $0.320$ with \texttt{GPT-5.2}, and from $0.089$ to $0.402$ with \texttt{DeepSeek-V3.2}.

\textbf{Distributional alignment.}
\textsc{MetaPersona} improves population-level fit on MIST, achieving the best Bias, W-dist, and KS across all three models. The ISSP results are mixed: \textsc{MetaPersona} has the lowest W-dist only with \texttt{DeepSeek-V3.2}; \textsc{LLM-Heuristic} is best with \texttt{GPT-5.2} and \texttt{Claude~Haiku~4.5}. Under \texttt{GPT-5.2}, \textsc{MetaPersona} does not lead on any reported ISSP metric. These differences are evaluated within each task; their absolute magnitudes are not compared across tasks because MIST and ISSP use different metric scales.

\textbf{Structural realism.}
A useful synthetic population must reproduce real-world \emph{dependencies} among attributes, not just their marginals. We measure the Pearson correlation $r$ along key variable paths in the ground-truth data and compare it to each method (Table~\ref{tab:structural-realism}), scoring on the absolute deviation $|r_{\text{method}} - r_{\text{GT}}|$. We pick one diagnostic path per task, each probing a different structure. For misinformation, cognitive reflection, measured by CRT, $\to$ misinformation discernment is a strong positive effect \citep{pennycook2019lazy} that demographics alone cannot express. For income redistribution, education level $\to$ support for reducing income differences is weak and non-monotonic \citep{attewell2022redistribution}; this path penalizes methods that impose an oversimplified stereotype. For AI-tool sentiment,
\textsf{seniority}~$\to$~sentiment is close to null, with favorability roughly flat across experience levels \citep{stackoverflow_survey}.Overall, \textsc{MetaPersona} is closest to the ground-truth correlation in two of the three diagnostic paths, reducing $|\Delta r|$ to $0.020$ for MIST and $0.036$ for SO. 

\begin{table}[t]
\centering
\caption{%
  Structural realism using \texttt{GPT-5.2}; results for other models are reported in Appendix~\ref{appendix:extended-results}. Pearson $r$ along one diagnostic path per task in the ground-truth data versus each method's prediction. Closest to ground truth in \textbf{bold}; $|\Delta r|$ is the absolute deviation from the ground-truth $r$ (lower is better).
}
\label{tab:structural-realism}
\footnotesize
\setlength{\tabcolsep}{4pt}
\begin{tabular}{ll cc}
\toprule
\textbf{Task / Path} & \textbf{Method} & $r$ & $|\Delta r|\,\downarrow$ \\
\midrule
\multirow{3}{*}{\shortstack[l]{MIST\\ CRT $\to$ discern.\\ (expect $+$)}}
  & Ground truth         & 0.313 & --- \\
  & LLM-Heuristic        & 0.238 & 0.075 \\
  & \textsc{MetaPersona} & \textbf{0.293} & \textbf{0.020} \\
\addlinespace
\multirow{5}{*}{\shortstack[l]{ISSP\\ education $\to$ support\\ (expect $\approx0$/mixed)}}
  & Ground truth         & 0.043 & --- \\
  & Demo-Only            & \textbf{0.107} & \textbf{0.064} \\
  & Backstory            & 0.147 & 0.104 \\
  & LLM-Heuristic        & 0.152 & 0.109 \\
  & \textsc{MetaPersona} & 0.123 & 0.080 \\
\addlinespace
\multirow{5}{*}{\shortstack[l]{SO\\ seniority $\to$ sentiment\\ (sign uncertain)}}
  & Ground truth         & 0.041 & --- \\
  & Demo-Only            & 0.128 & 0.087 \\
  & Backstory            & 0.105 & 0.064 \\
  & LLM-Heuristic        & 0.175 & 0.134 \\
  & \textsc{MetaPersona} & \textbf{0.077} & \textbf{0.036} \\
\bottomrule
\end{tabular}
\begin{minipage}{0.95\linewidth}
\scriptsize
\vspace{1pt}
\textit{Note.} For the misinformation path, CRT (cognitive reflection) is an inferred latent attribute rather than a demographic field, so only \textsc{LLM-Heuristic} and \textsc{MetaPersona} explicitly encode it.
\vspace{-0.1in}
\end{minipage}
\end{table}

\section{Application: \textsc{MetaPersona-Studio}}
\label{sec:metapersona-studio}
\vspace{-0.05in}

To explore how the framework could be used without code, we developed a work-in-progress interactive prototype, \textsc{MetaPersona-Studio}. 
A researcher specifies a task query and selects a grounding population, and the Studio retrieves task-relevant literature, builds the persona graph, surfaces the selected latent attributes, and generates a synthetic population, exposing demographic filters (e.g., country, age, education, employment) so the base population can be customized to the study at hand. The interface returns the generated personas and graph for inspection and export.

\section{Discussion and Conclusion}
\vspace{-0.05in}

We presented a pipeline that turns scattered empirical evidence into task-grounded synthetic populations: \textsc{MetaPersona-DB} distills 11{,}000+ human-subjects studies into variable-level structure, \textsc{MetaPersona} uses that evidence to retrieve task-relevant attributes and sample populations under literature-derived priors, and the \textsc{MetaPersona-Studio} prototype explores how this workflow can be made accessible to interdisciplinary users.

\textbf{When does task-grounded structure help?}
Our results suggest the gains are largest when an outcome depends on latent attributes that demographics alone cannot capture. This is clearest on misinformation belief and AI-tool sentiment: on Stack Overflow with \texttt{GPT-5.2}, quadratic-weighted kappa rises from the best baseline value
of $0.073$ to $0.320$, indicating that empirically selected latents recover meaningful differences between people rather than merely matching aggregate marginals. Where a relationship is weak, as for education $\to$ redistribution support on ISSP where all methods converge, \textsc{MetaPersona} offers no advantage.

\textbf{Conclusion.}
Faithful synthetic populations require not only richer descriptions but the right attributes and their empirical dependencies. By grounding both in measured social science evidence, \textsc{MetaPersona} moves persona generation from a heuristic step toward a principled one, and puts it in the hands of the communities best positioned to use it.

\newpage
\section*{Limitations and Future Work}
\vspace{-0.1in}

Our evidence base is bounded in coverage: we index six social science disciplines rather than all JCR fields, and we draw on journal articles only, omitting archival venues such as HCI conferences that also report human-subjects studies. We extract variables from titles and abstracts rather than full text. This is partly by design, since abstracts surface the most important variables and headline relationships while full-text mining introduces noise from robustness checks and incidental measures, but it biases the evidence base toward primary and well-powered effects. Coverage is further shaped by access since many studies are not open access. Natural next steps are to broaden the corpus to additional disciplines and conference venues, label the open-access subsample, and extend extraction to full text where licensing permits. A second boundary concerns the demographic substrate: the current grounding assumes a sampling frame defined by observable demographic attributes, which may not be available for all target populations of interest. Two further directions follow: validating the census-microdata substrate (\S\ref{subsec:demo-sampling}) on non-benchmarked tasks, and extending the dependency graph toward richer structures that capture moderation and context-dependent effects.

\section*{Ethical considerations}
\vspace{-0.1in}

\textsc{MetaPersona} is intended to support, not replace, human-subjects research. Even when empirically grounded, simulated responses reflect aggregate patterns rather than the lived experience of real people, and they should not be treated as evidence about actual individuals or used to substitute for engagement with affected communities on consequential decisions. As with any method that models human populations, care is needed
when applying it to groups for which the underlying evidence is limited. We also note dual-use potential: realistic synthetic audiences could be used to optimize persuasive messaging, and we intend \textsc{MetaPersona-Studio} for research and transparent analysis. Finally, we used ChatGPT exclusively to improve minor grammar in the final manuscript text.

\section*{Broader Impacts}
\vspace{-0.1in}

By reducing persona construction cost to under \$0.5 per task and making it flat in population size, \textsc{MetaPersona} removes a practical barrier to large-scale social simulation. We note, however, that persona generation is not the dominant cost in most simulation pipelines; the simulation itself typically dominates. \textsc{MetaPersona}'s contribution is therefore to make the persona layer both cheaper and more rigorous, so researchers can invest remaining budget in simulation fidelity rather than population construction. Grounded populations support a range of applications: synthetic audiences for pre-screening messaging in marketing and consumer research \citep{sarstedt2024silicon}, and hypothesis screening and opinion forecasting in computational social science \citep{horton2023homo, manning2024automated}. Simulated populations should complement rather than displace human data, a point we develop in the Ethical Considerations.


\section*{Code and Demo Availability}
This manuscript describes work in progress. The research code and public
\textsc{MetaPersona-Studio} demo are being prepared for release and will be
made publicly available with the next version of the manuscript.


\bibliography{custom}

\newpage
\appendix

\section{Extended Related Work}
\label{appendix:extended-related-work}

\paragraph{Limits of Naive Persona Generation}
Recent work converges on a common diagnosis: naive persona generation often adds spurious structure rather than improving realism. Increasing the amount of LLM-generated persona content can amplify bias and push simulated populations away from empirical targets \citep{li2026promise}, while large-scale benchmarks show that synthetic social-science data tend to collapse toward typical profiles, exaggerate associations among variables, and fail to reproduce more complex life-course patterns \citep{xie2026evaluating}. These failures suggest that persona realism is not simply a matter of adding richer descriptions or scaling to stronger models. A second line of evidence shows why fixed or freely generated personas are especially fragile: they often encode the wrong sources of variation. LLM simulations can misportray and flatten identity groups, reducing within-group diversity even when group membership is explicitly specified \citep{wang2025flatten, yu2025large}. More generally, persona prompting helps only when the included attributes are predictive of the target outcome; otherwise, persona variables explain little behavioral variance \citep{hu2024quantifying}. Together, these findings motivate our central premise: the key problem is not to generate more detailed personas, but to identify task-relevant attributes and preserve their empirical dependencies.

\paragraph{Constructing More Realistic Personas}
Existing work on persona construction can be grouped into three main approaches. First, demographic and census-based methods sample personas to match population marginals or known demographic distributions, improving coverage of observable groups but usually relying on fixed schemas \citep{argyle2023one, ge2024scaling, yang2024oasis, piao2025agentsociety}. Second, text-grounded methods seek realism through richer individual evidence: interview-based agents provide strong individual fidelity but are costly to collect \citep{park2024generative}, while backstory and narrative-persona methods improve coherence and plausibility through generated life histories or persona-conditioned dialogue \citep{moon2024anthology}. Third, post-hoc alignment methods use resampling or transport to match a fixed persona pool to a target attribute distribution. \citet{hu2025population} is closest to our setting, but aligns a preselected Big Five schema rather than deciding which attributes a task requires. Across these approaches, the schema-selection problem remains largely unaddressed. We note that \citet{li2026promise} use the term ``meta-persona'' to describe aggregate demographic profiles for LLM steering; our naming is inspired by their framing, extending it toward a meta-level empirical evidence base drawn from the social science literature.

\paragraph{Adapting the Simulator Rather than the Persona.}
A separate line of work improves social simulation by adapting the simulator rather than the persona. SOCSCI-style fine-tuning trains models directly on large collections of human-subjects responses, improving alignment with empirical response distributions across unseen social-science experiments \citep{kolluri2025finetuning}. Related benchmark work also suggests that domain-specific training can improve population-level statistical realism \citep{xie2026evaluating}. These approaches condition behavioral regularities into model weights, whereas \textsc{MetaPersona} structures the input population by selecting task-relevant variables and dependencies. The two directions are therefore complementary: a response-tuned simulator could be combined with \textsc{MetaPersona}-structured personas. More broadly, recent theory-grounded accounts argue that faithful social simulation requires cognitive and social structure beyond demographic labels alone \citep{li2026simulating, ng2026social}, a view our literature-grounded dependency graphs operationalize.

\paragraph{Evaluating Population-Level Realism}
A final line of work asks how synthetic populations should be evaluated. Early studies often focus on individual accuracy, testing whether a simulated agent predicts a specific person's response \citep{park2024generative}; however, this metric can reward degenerate mean-prediction behavior without requiring a model to represent the population. Recent work therefore emphasizes distributional alignment, using metrics such as Wasserstein distance, KS distance, Cramér's V, and moment errors to compare simulated and empirical response distributions, including on public-opinion benchmarks such as \textsc{OpinionQA} and \textsc{SubPOP} \citep{santurkar2023whose, suh2025language}. More stringent frameworks further require structural realism: synthetic populations should preserve bivariate associations, multivariate predictive relationships, and life-course sequence patterns, not merely marginal distributions \citep{xie2026evaluating}. Our evaluation follows this view by measuring \textsc{MetaPersona} across individual accuracy, distributional alignment, and structural realism.


\section{Cost Analysis}
\label{appendix:cost-analysis}

We evaluate the cost of generating the persona population for each case study. Costs are measured on the samples in Section~\ref{sec:case_studies}, with GPT-5.2 used as the generation model. 

\begin{table}[h]
\centering
\footnotesize 
\renewcommand{\arraystretch}{1.2}
\setlength{\tabcolsep}{4.5pt}
\caption{Persona construction cost across case studies.}
\label{tab:generation_cost}

\begin{tabular}{@{} l c c c @{}}
\toprule
\midrule
\textbf{Method} & \textbf{MIST} & \textbf{ISSP} & \textbf{SO} \\
\midrule
Number of personas & 1000 & 2915 & 2000 \\
\midrule
LLM-Heuristic & $\sim$\$2.27 & $\sim$\$7.91 & $\sim$\$4.64 \\
Backstory & $\sim$\$8.52 & $\sim$\$27.68 & $\sim$\$16.84 \\
\textsc{MetaPersona} & $\sim$\$0.16 & $\sim$\$0.25 & $\sim$\$0.18 \\
\bottomrule
\end{tabular}
\end{table}

Table~\ref{tab:generation_cost} summarizes costs across methods and tasks. \textsc{MetaPersona} has lower persona-construction cost than LLM-Heuristic and Backstory because it does not generate a new profile for each sampled individual. More importantly, this gap widens as the target population grows. LLM-Heuristic and Backstory require one or more LLM calls per persona, so their construction costs increase with the number of simulated individuals. \textsc{MetaPersona} builds the task-specific persona graph once and then samples individuals from that graph, adding statistical sampling operations rather than new LLM generations.

\section{Dataset Statistics, Coverage, and Design Choices}
\label{appendix:dataset}

\begin{table*}[t]
\centering
\small
\begin{tabular}{
>{\raggedright\arraybackslash}p{0.28\linewidth}
>{\raggedright\arraybackslash}p{0.36\linewidth}
>{\raggedright\arraybackslash}p{0.12\linewidth}
}
\toprule
\textbf{Validation component} & \textbf{Measure} & \textbf{Result} \\
\midrule

\multicolumn{3}{>{\raggedright\arraybackslash}p{0.86\linewidth}}{\textit{Validation 1: Empirical paper labeling}} \\
Human agreement & Cohen's $\kappa$ & 0.94 \\
Human agreement & Raw agreement & 97\% \\
LLM vs. adjudicated ground truth & Precision & 86\% \\
LLM vs. adjudicated ground truth & Recall & 97\% \\
LLM vs. adjudicated ground truth & F1 & 91\% \\

\midrule
\multicolumn{3}{>{\raggedright\arraybackslash}p{0.86\linewidth}}{\textit{Validation 2: Full-text PDF annotation validation}} \\
Human agreement & Cohen's $\kappa$ & 0.86 \\
Human agreement & Raw agreement & 91\% \\
Variables & Supported or partially supported & 94\% \\
Measurement forms & Supported or partially supported & 90\% \\
Distribution priors & Plausible or partially plausible & 88\% \\
Paths & Supported or partially supported & 91\% \\
Directions & Correct & 93\% \\
Mediators/moderators & Correct or partially correct & 86\% \\
Overall annotations & Usable for graph construction & 91\% \\
\bottomrule
\end{tabular}
\caption{Human validation results. Validation 1 measures empirical-paper labeling reliability and LLM classification accuracy against adjudicated human labels. Validation 2 measures whether extracted annotations are supported by full-text PDFs.}
\label{tab:human-validation-results}
\end{table*}

\subsection{Annotation Taxonomy}

\textbf{Individual-level variable.}
We define an \textit{individual-level variable} as any measurable attribute, belief, disposition, behavior, or outcome that can vary across people and can be
assigned to a single persona. This includes relatively stable traits, such as age, education, ideology, or risk aversion, as well as task-specific states or outcomes, such as trust in government, vaccination intention, misinformation
discernment, or AI tool adoption. We exclude purely study-level descriptors, such as country, institution, policy setting, or sampling frame, unless they are used as person-level attributes or as moderators of individual-level relationships.

Table~\ref{tab:annotation-taxonomy} summarizes the annotation schema used to extract individual-level variables and directed variable relationships from paper titles and abstracts. The schema contains three groups of annotations:
study-level descriptors, variable-level annotations, and path-level annotations.

\subsection{Dataset Statistics}

\begin{figure}[h]
  \centering
  \includegraphics[width=\linewidth]{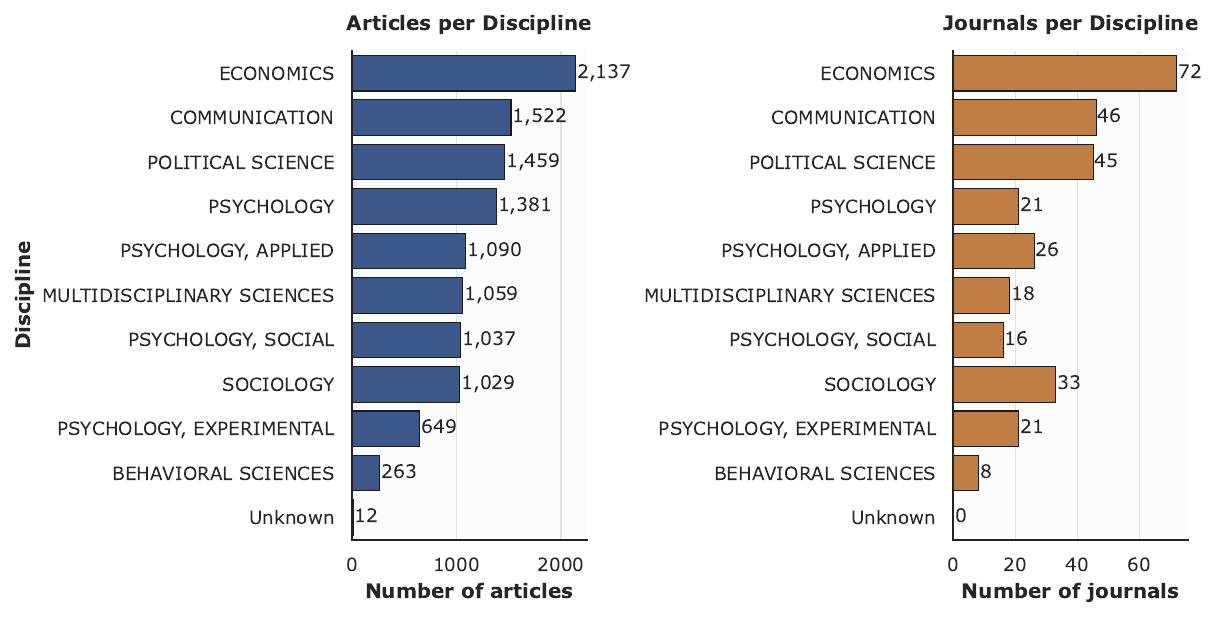}
  \caption{Distribution of articles in \textsc{MetaPersona-DB} across disciplines and journals.}
  \label{fig:dataset_statistics}
\end{figure}

Figure~\ref{fig:dataset_statistics} shows the distribution of articles across academic disciplines and publication venues, illustrating the disciplinary breadth of the corpus used for variable and path extraction.

\subsection{Human Validation}
\label{appendix:human-validation}

We conducted two human validation studies. The first evaluates whether papers included in \textsc{MetaPersona-DB} are empirical social-science studies suitable for variable and path extraction. The second evaluates whether the LLM-extracted variables, distributions, and paths are supported by the full-text PDFs. Each validation task was performed by two independent annotators on a sample of 100 papers.

\paragraph{Validation 1: Empirical Paper Labeling}

This validation assesses the task validity and automatic labeling accuracy of the empirical-paper selection step. We randomly sample 100 papers from the candidate corpus. For each paper, two annotators independently label the paper as empirical or non-empirical, using the same LLM classification prompt in Appendix~\ref{appendix:prompt} as the labeling guideline. We compute inter-annotator agreement to estimate the reliability of this labeling task. The two annotators then discuss disagreements and produce an adjudicated ground-truth label for each paper. We evaluate the LLM-generated labels against this ground truth using precision, recall, and F1. Thus, human agreement measures task validity, while LLM F1 measures the accuracy of the automatic classification pipeline.

\paragraph{Validation 2: Full-text PDF Annotation Validation}

This validation assesses whether the LLM-extracted annotations are supported by the full-text papers. We randomly sample 100 open access papers with available PDFs from the extraction output. For each paper, two annotators independently review the PDF and the corresponding LLM annotations, including extracted variables, measurement forms, distribution priors, paths, directions, mediators, and moderators.

Annotators label each annotation component as \textit{supported}, \textit{partially supported}, or \textit{unsupported}. Variable-level validation checks whether the variable is present, assigned the correct role, and measured as annotated. Path-level validation checks whether the source and target variables are correct, whether the direction and relationship type match the paper, and whether any mediator or moderator is correctly identified. Distribution priors are evaluated as coarse construct-level judgments unless the paper reports explicit descriptive statistics.

We calculate inter-annotator agreement, resolve disagreements through discussion, and use the adjudicated labels to estimate the support rate of the LLM annotations against full text.

\paragraph{Results.}

Table~\ref{tab:human-validation-results} summarizes the results of both validation studies. For empirical-paper labeling, annotators achieved very high agreement (Cohen's $\kappa = 0.94$, raw agreement $=97\%$), indicating that identifying empirical papers is a reliable and low-ambiguity task. Compared against adjudicated human labels, the LLM paper-selection step achieved high recall ($97\%$), precision ($86\%$), and F1 ($91\%$), suggesting that the classifier is conservative and rarely excludes empirical studies, at the cost of a small number of false positives.

For full-text PDF validation, annotators also achieved strong agreement (Cohen's $\kappa = 0.86$, raw agreement $=91\%$), indicating that support judgments are sufficiently reliable despite the greater complexity of the task. Most extracted annotations were supported or partially supported by the full text, including variables ($94\%$), measurement forms ($90\%$), paths ($91\%$), and relationship directions ($93\%$). Distribution priors achieved slightly lower support ($88\%$), consistent with their interpretation as construct-level judgments rather than study-specific estimates, while mediator and moderator annotations were somewhat harder to validate ($86\%$). Overall, $91\%$ of annotations were judged usable for persona-graph construction.

\begin{table*}[t]
\centering
\footnotesize
\begin{tabular}{p{0.20\linewidth} p{0.3\linewidth} p{0.4\linewidth}}
\toprule
\textbf{Field} & \textbf{Definition} & \textbf{Examples / Allowed Values} \\
\midrule

\multicolumn{3}{l}{\textit{Study-level descriptors}} \\
\midrule

Domain or subfield &
The specific empirical or disciplinary area of the study. &
Political communication; health psychology; labor economics; organizational behavior. \\

Agent types &
The social actors involved in the study. &
Voters; consumers; patients; employees; managers; firms; institutions. \\

Context: country &
The geographic setting of the study. &
USA; Germany; China; cross-national sample. \\

Context: population &
The sample, population, or empirical setting studied. &
Registered voters; undergraduate students; low-income households; Fortune 500 employees. \\

Main finding &
A one-sentence summary of the paper's primary empirical result. &
``Economic insecurity is associated with lower institutional trust.'' \\

Theoretical constructs &
Named theories, concepts, or conceptual frameworks used by the paper. &
Social identity; political polarization; institutional trust; risk perception; loss aversion. \\

\midrule
\multicolumn{3}{l}{\textit{Variable-level annotations}} \\
\midrule

Independent variables &
Predictors, treatments, policies, exposures, or upstream variables. &
Campaign exposure; income inequality; social media use; economic insecurity. \\

Dependent variables &
Outcomes, behaviors, attitudes, beliefs, or downstream variables. &
Voter turnout; policy support; institutional trust; life satisfaction. \\

Mechanisms or mediators &
Variables or processes explaining why an upstream variable affects a downstream variable. &
Perceived threat; risk perception; social comparison; identity salience. \\

Moderators &
Variables that condition the strength, direction, or presence of a relationship. &
Political ideology; gender; prior belief; age; social context. \\

Persona traits or attributes &
Individual-level attributes useful for constructing agent personas. &
Age; gender; education; income; ideology; trust in government; risk aversion; media use. \\

Distribution shape &
A coarse prior over the population-level shape of a persona-relevant variable. &
Bell-shaped; right-skewed; left-skewed; bimodal; uniform; categorical; sparse/zero-inflated; unclear. \\

Measurement form &
How the variable is measured or represented when stated or inferable. &
Scale score; binary label; categorical label; behavioral count; frequency; index; open-ended response; unclear. \\

\midrule
\multicolumn{3}{l}{\textit{Path-level annotations}} \\
\midrule

Source variable &
The predictor, treatment, exposure, or upstream variable in a directed path. &
Economic insecurity in ``economic insecurity $\rightarrow$ political distrust.'' \\

Target variable &
The downstream variable affected by or associated with the source. &
Political distrust in ``economic insecurity $\rightarrow$ political distrust.'' \\

Relationship type &
The type of variable relationship reported or strongly implied. &
Direct effect; mediated effect; moderated effect; correlation; association; mechanism; unclear. \\

Direction &
The signed direction of the relationship. &
Positive; negative; mixed; null; unclear. \\

Mediator &
The intervening variable in a mediated relationship, if applicable. &
Perceived threat in ``economic insecurity $\rightarrow$ perceived threat $\rightarrow$ anti-immigration attitudes.'' \\

Moderator &
The subgroup, condition, or context that changes the relationship, if applicable. &
Political interest in ``campaign exposure has a stronger effect among politically interested voters.'' \\

Evidence text &
A short phrase from the abstract supporting the extracted path. &
``is associated with lower life satisfaction''; ``increases perceived threat.'' \\

Confidence &
The annotator's confidence that the path is supported by the abstract. &
High; medium; low. \\

Path summary &
A concise natural-language summary of the main relationship structure. &
``Economic insecurity increases perceived threat, which predicts anti-immigration attitudes.'' \\

\bottomrule
\end{tabular}
\caption{Annotation taxonomy for extracting persona-relevant variables and variable relationships from paper metadata.}
\label{tab:annotation-taxonomy}
\end{table*}

\newpage
\clearpage
\section{Metric Definitions}
\label{appendix:experiment-design}

All metrics are computed on matched persona--respondent pairs after dropping
rows with missing predictions or out-of-range codes. We describe each metric
by task type below.

\subsection*{Case Study 1: Misinformation Belief (MIST)}

Raw MIST-20 scores (integer, $0$--$20$) are normalized to $[0,1]$ by dividing
by 20 before computing all numeric metrics. Let $y_i$ denote the normalized
ground-truth score and $\hat{y}_i$ the normalized predicted score for
respondent $i$, over $n$ matched pairs.

\paragraph{Root Mean Squared Error (RMSE).}
\[
  \mathrm{RMSE} = \sqrt{\frac{1}{n}\sum_{i=1}^{n}(y_i - \hat{y}_i)^2}
\]

\paragraph{Mean Absolute Error (MAE).}
\[
  \mathrm{MAE} = \frac{1}{n}\sum_{i=1}^{n}|y_i - \hat{y}_i|
\]

\paragraph{Bias (Mean Signed Error).}
\[
  \mathrm{Bias} = \frac{1}{n}\sum_{i=1}^{n}(\hat{y}_i - y_i)
\]
Positive bias indicates systematic over-prediction; negative bias indicates
under-prediction.

\paragraph{Wasserstein Distance.}
The 1-Wasserstein (earth-mover) distance between the empirical distributions
of $\{y_i\}$ and $\{\hat{y}_i\}$, computed via \texttt{scipy.stats.wasserstein\_distance}.

\paragraph{Kolmogorov--Smirnov Statistic (KS).}
The two-sample KS statistic $D = \sup_t |F_n(t) - \hat{F}_n(t)|$, where
$F_n$ and $\hat{F}_n$ are the empirical CDFs of the ground-truth and
predicted score distributions, computed via \texttt{scipy.stats.ks\_2samp}.

\subsection*{Case Studies 2 and 3: Ordinal Outcomes (ISSP and Stack Overflow)}

Both tasks predict 5-point ordinal items. For Case Study 2 we predict
\texttt{GOV1} (1 = Strongly agree, 5 = Strongly disagree). For Case Study 3
we predict \texttt{AISent} and \texttt{AISelect}, each coded 1--5; responses
coded 6 (\emph{Unsure}) in \texttt{AISent} are excluded before metric
computation.

Metrics for ordinal outcomes are computed from their integer response codes.
For ISSP, MAE and Wasserstein distance are reported on the original 1--5
scale. Within-one accuracy records whether a prediction is within one ordinal
step of the ground truth, and QWK is computed on the raw integer codes. Let
$r_i \in \{v_{\min},\ldots,v_{\max}\}$ be the ground-truth code and
$\hat{r}_i$ the predicted code. Because metric scales differ across tasks,
their absolute magnitudes should only be compared within a task.

\paragraph{Mean Absolute Error (MAE).}
For ISSP, MAE is computed on the original ordinal codes:
$\mathrm{MAE} = \frac{1}{n}\sum_i |r_i - \hat{r}_i|$.

\paragraph{Within-One Accuracy (W-1 Acc).}
Proportion of predictions within one ordinal step of the ground truth:
\[
  \text{W-1 Acc} = \frac{1}{n}\sum_{i=1}^{n}
    \mathbf{1}\!\left[|r_i - \hat{r}_i| \leq 1\right]
\]

\paragraph{Quadratic Weighted Kappa (QWK).}
Computed on raw integer codes over the full confusion matrix $O$, with
quadratic weight matrix $w_{ij} = (i-j)^2 / (K-1)^2$ where
$K = v_{\max} - v_{\min} + 1$:
\[
  \mathrm{QWK} = 1 - \frac{\sum_{i,j} w_{ij}\, O_{ij}}
                           {\sum_{i,j} w_{ij}\, E_{ij}}
\]
where $E_{ij}$ is the expected count under independence.
QWK $= 1$ indicates perfect agreement; QWK $= 0$ indicates chance-level agreement.

\paragraph{Wasserstein Distance.}
Computed on raw ordinal codes between ground-truth and predicted
distributions, as for Case Study 2. For Case Study 3, the Wasserstein
distance is macro-averaged over \texttt{AISent} and \texttt{AISelect}.

\section{Extended Experiment Results}
\label{appendix:extended-results}

\begin{table*}[t]
\centering
\caption{
Structural realism for additional models. Pearson $r$ along one diagnostic path per task in the ground-truth data versus each method's prediction. Closest to ground truth within each model is shown in \textbf{bold}; $|\Delta r|$ is the absolute deviation from the ground-truth $r$ (lower is better).
}
\label{tab:structural-realism-extended}
\footnotesize
\setlength{\tabcolsep}{4pt}
\begin{tabular}{lllcc}
\toprule
\textbf{Task / Path} & \textbf{Model} & \textbf{Method} & $r$ & $|\Delta r|\,\downarrow$ \\
\midrule

\multirow{7}{*}{\shortstack[l]{MIST\\ CRT $\to$ discern.\\ (expect $+$)}}
& \multirow{3}{*}{Claude Haiku 4.5}
  & Ground truth         & 0.313 & --- \\
& & LLM-Heuristic        & -0.172 & 0.485 \\
& & \textsc{MetaPersona} & \textbf{-0.021} & \textbf{0.334} \\

& \multirow{3}{*}{DeepSeek v3.2}
  & Ground truth         & 0.313 & --- \\
& & LLM-Heuristic        & 0.423 & 0.110 \\
& & \textsc{MetaPersona} & \textbf{0.052} & \textbf{0.261} \\

\addlinespace
\multirow{11}{*}{\shortstack[l]{ISSP\\ education $\to$ support\\ (expect $\approx0$/mixed)}}
& \multirow{5}{*}{Claude Haiku 4.5}
  & Ground truth         & 0.043 & --- \\
& & Demo-Only            & 0.188 & 0.145 \\
& & Backstory            & \textbf{0.126} & \textbf{0.083} \\
& & LLM-Heuristic        & 0.188 & 0.145 \\
& & \textsc{MetaPersona} & 0.143 & 0.100 \\

& \multirow{5}{*}{DeepSeek v3.2}
  & Ground truth         & 0.043 & --- \\
& & Demo-Only            & \textbf{0.151} & \textbf{0.108} \\
& & Backstory            & 0.172 & 0.129 \\
& & LLM-Heuristic        & 0.181 & 0.138 \\
& & \textsc{MetaPersona} & 0.154 & 0.111 \\

\addlinespace
\multirow{11}{*}{\shortstack[l]{SO\\ seniority $\to$ sentiment\\ (sign uncertain)}}
& \multirow{5}{*}{Claude Haiku 4.5}
  & Ground truth         & 0.041 & --- \\
& & Demo-Only            & 0.455 & 0.414 \\
& & Backstory            & 0.403 & 0.362 \\
& & LLM-Heuristic        & 0.388 & 0.347 \\
& & \textsc{MetaPersona} & \textbf{0.075} & \textbf{0.034} \\

& \multirow{5}{*}{DeepSeek v3.2}
  & Ground truth         & 0.041 & --- \\
& & Demo-Only            & 0.582 & 0.541 \\
& & Backstory            & 0.603 & 0.562 \\
& & LLM-Heuristic        & 0.323 & 0.282 \\
& & \textsc{MetaPersona} & \textbf{0.134} & \textbf{0.093} \\

\bottomrule
\end{tabular}

\begin{minipage}{0.98\linewidth}
\scriptsize
\vspace{2pt}
\textit{Note.} For the misinformation path, CRT (cognitive reflection) is an inferred latent attribute rather than a demographic field, so only \textsc{LLM-Heuristic} and \textsc{MetaPersona} explicitly encode it. For ISSP and Stack Overflow, Demo-Only, Backstory, and LLM-Heuristic correspond to Baseline~1, Baseline~2, and Baseline~3 respectively.
\end{minipage}
\end{table*}

\paragraph{Additional model results on structural realism.}

Table~\ref{tab:structural-realism-extended} reports the structural-realism analysis for additional foundation models. Consistent with the main results, \textsc{MetaPersona} generally produces variable relationships that are closer to the empirical data than prompting-based baselines, particularly on the Stack Overflow task where the ground-truth relationship is weak and easy to exaggerate. Across models, \textsc{MetaPersona} tends to recover more conservative and empirically plausible correlations, whereas baseline prompting methods often overstate relationship strength.

At the same time, performance varies across models and tasks. In some settings, especially when the expected empirical relationship is weak or ambiguous, heuristic baselines can occasionally achieve comparable or lower absolute deviation from the ground truth. We therefore view the appendix results as supporting the broader claim that task-grounded latent structure improves robustness across model families, while also highlighting that structural realism depends in part on the capabilities and inductive biases of the underlying LLM.

\section{Prompts}
\label{appendix:prompt}

\textbf{Prompt 1: Empirical Paper Classification}

\begin{promptbox}
You are a careful annotator. Given the metadata for one academic paper, decide whether it is an EMPIRICAL HUMAN-INVOLVING SOCIAL SCIENCE STUDY.

\medskip
For this task, EMPIRICAL HUMAN-INVOLVING SOCIAL SCIENCE STUDY includes BOTH:\\
1. INDIVIDUAL\_STUDY: an original empirical study analyzing real-world human data\\
2. SYSTEMATIC\_REVIEW: a systematic review, scoping review, or meta-analysis that synthesizes empirical human studies on individual-level attributes and attitudinal or behavioral outcomes

\medskip
In other words:\\
- If a paper is an eligible original empirical human study, label it as EMPIRICAL.\\
- If a paper is an eligible systematic review/meta-analysis of such human studies, also label it as EMPIRICAL.\\
- Then separately classify whether it is INDIVIDUAL\_STUDY or SYSTEMATIC\_REVIEW.

\medskip
Inclusion criteria for label = "EMPIRICAL":\\
Label as "EMPIRICAL" if BOTH of the following are true:\\
1. The paper is about humans and focuses on the relationship between individual-level attributes (e.g., demographics, preferences, beliefs, values, personality, risk attitudes, identities, experiences, prior exposures) and individual-level attitudinal or behavioral outcomes.\\
2. The paper is either:\\
\hspace*{1em}a. an original empirical study using real-world human data (e.g., surveys, experiments, field experiments, administrative data, census data, microdata, voter data, household data, interviews, qualitative data), OR\\
\hspace*{1em}b. a systematic review, scoping review, or meta-analysis that explicitly synthesizes empirical human studies on such relationships.

\medskip
Exclusion criteria for label = "NOT\_EMPIRICAL":\\
Label as "NOT\_EMPIRICAL" if ANY are true:\\
- Pure theory, mathematical modeling, or simulations without real human data.\\
- Methods papers (econometrics, algorithms, statistics) without applied human data.\\
- Studies about natural systems (climate, physics, biology) without humans.\\
- Conceptual or narrative policy/literature reviews that summarize prior work without a systematic evidence synthesis.\\
- Reviews that are not about individual-level human attributes and attitudinal or behavioral outcomes.\\
- Studies that only measure physiological, neurological, or biomechanical responses without attitudes, decisions, intentions, or observable social behaviors.\\
- Studies where human data are used only as controls, descriptive statistics, or intermediate variables, rather than as the main object of analysis.

\medskip
Paper type classification:\\
- "INDIVIDUAL\_STUDY": original empirical study using a specific human sample or dataset\\
- "SYSTEMATIC\_REVIEW": systematic review, scoping review, evidence synthesis, or meta-analysis of prior empirical human studies\\
- "OTHER": not an individual study or systematic review\\
- "UNCERTAIN": cannot tell from the metadata

\medskip
Important annotation rules:\\
- A paper can be:\\
\hspace*{1em}- label = "EMPIRICAL", paper\_type = "INDIVIDUAL\_STUDY"\\
\hspace*{1em}- label = "EMPIRICAL", paper\_type = "SYSTEMATIC\_REVIEW"\\
- Narrative reviews should usually be:\\
\hspace*{1em}- label = "NOT\_EMPIRICAL", paper\_type = "OTHER"\\
- Meta-analyses and systematic reviews count as EMPIRICAL for this task if they systematically synthesize empirical human studies within scope.\\
- If the metadata are insufficient, use "UNCERTAIN".

\medskip
Output strictly as compact JSON:
\begin{quote}
\ttfamily\scriptsize
\{\\
\hspace*{1em}"label": "<EMPIRICAL|NOT\_EMPIRICAL|UNCERTAIN>",\\
\hspace*{1em}"paper\_type": "<INDIVIDUAL\_STUDY|SYSTEMATIC\_REVIEW|OTHER|UNCERTAIN>",\\
\hspace*{1em}"rationale": "<one short sentence>",\\
\hspace*{1em}"confidence": <float 0-1>\\
\}
\end{quote}

\medskip
Now classify this paper:

\medskip
Title: \{title\}\\
Abstract: \{abstract\}\\
Journal: \{journal\}\\
Authors: \{authors\}
\end{promptbox}

\noindent\textbf{Prompt 2: Variable, Distribution, and Path Extraction}

\begin{promptbox}
You are an expert academic annotator in the social sciences. Given the metadata for one academic study, extract persona-relevant variables and reported relationships.

\medskip
Input:\\
Title: \{title\}\\
Abstract: \{abstract\}\\
Journal: \{journal\}\\
Authors: \{authors\}

\medskip
Return a strict JSON object with the following fields.

\medskip
1. \textbf{domain\_or\_subfield}: specific academic subfield, such as political communication, health psychology, labor economics, or organizational behavior.

\medskip
2. \textbf{agent\_types}: actors involved in the study, such as voters, consumers, patients, employees, managers, or firms.

\medskip
3. \textbf{context}: country or geographic setting, and the sample or study population.

\medskip
4. \textbf{main\_finding}: one concise sentence summarizing the primary result.

\medskip
5. \textbf{theoretical\_constructs}: key concepts, theories, or frameworks used in the paper.

\medskip
6. \textbf{variable\_annotations}: list of individual-level variables. Use the following allowed values:
\texttt{role}: IV, DV, mediator, moderator, trait, unclear;
\texttt{type}: demographic, psychographic, behavioral, socioeconomic, cultural, contextual, unclear;
\texttt{measure}: scale, binary, categorical, count, frequency, index, open-ended, unclear;
\texttt{shape}: bell, right-skewed, left-skewed, bimodal, uniform, categorical, sparse, unclear.

\begin{quote}
\ttfamily\scriptsize
\{\\
\hspace*{1em}"name": "String",\\
\hspace*{1em}"role": "IV|DV|mediator|moderator|trait|unclear",\\
\hspace*{1em}"type": "demographic|psychographic|behavioral|...",\\
\hspace*{1em}"measure": "scale|binary|categorical|count|...",\\
\hspace*{1em}"shape": "bell|right-skewed|bimodal|sparse|...",\\
\hspace*{1em}"shape\_rationale": "Short explanation",\\
\hspace*{1em}"confidence": "high|medium|low"\\
\}
\end{quote}

\medskip
7. \textbf{variable\_paths}: directed relationships reported or strongly implied by the title and abstract. Use the following allowed values:
\texttt{type}: direct, mediated, moderated, correlation, association, mechanism, unclear;
\texttt{direction}: positive, negative, mixed, null, unclear.

\begin{quote}
\ttfamily\scriptsize
\{\\
\hspace*{1em}"source": "String",\\
\hspace*{1em}"target": "String",\\
\hspace*{1em}"type": "direct|mediated|moderated|correlation|association|...",\\
\hspace*{1em}"direction": "positive|negative|mixed|null|unclear",\\
\hspace*{1em}"mediator": "String or null",\\
\hspace*{1em}"moderator": "String or null",\\
\hspace*{1em}"evidence": "Short supporting phrase",\\
\hspace*{1em}"confidence": "high|medium|low"\\
\}
\end{quote}

\medskip
8. \textbf{path\_summary}: one sentence summarizing the main relationship structure.

\medskip
Rules:\\
- Use only information reported or strongly implied in the title and abstract.\\
- Do not invent unsupported variables or paths.\\
- Prefer the paper's own terminology.\\
- Distinguish mediators from moderators.\\
- If a mechanism is reported, decompose it into directed edges.\\
- If an effect varies by subgroup or context, encode that variable as a moderator.\\
- If no relationship is reported, return an empty path list.\\
- If a field is not applicable or not mentioned, return null or [].

\medskip
Output valid JSON only. Do not include markdown or explanations.
\end{promptbox}

\noindent\textbf{Prompt 3: Backstory Persona Generation (Stack Overflow Case Study)}

\begin{promptbox}
System: Write a natural first-person professional life narrative that is consistent with the provided Description. Return only the backstory text. Do not mention that you are a persona, respondent, survey participant, language model, or AI. Do not simply repeat the description as a list.

\medskip
The narrative may include education, career path, developer role, work environment, industry context, learning habits, compensation context, and day-to-day engineering responsibilities, but must not answer any Stack Overflow survey question directly.

\medskip
User:\\
Below you will be asked to provide a short description of your background, and then answer a question.

\medskip
Description: \{description\}

\medskip
Question: Tell me about yourself. Please describe in detail.

\medskip
Answer: []
\end{promptbox}

\noindent\textbf{Prompt 4: LLM-Heuristic Persona Attribute Selection}

\begin{promptbox}
System: You are a research assistant that returns strict JSON.

\medskip
User:\\
I am building a persona schema for an LLM-based social simulation.

\medskip
Task: \{task\}\\
Population: \{population\}\\
Outcome: \{outcome\}

\medskip
Which individual-level attributes should be included in a persona so the simulation can realistically capture variation in this outcome?

\medskip
Return exactly the top \{top\_k\} most relevant person-level attributes.

\medskip
For each attribute, provide:\\
- \texttt{attribute}: short attribute name\\
- \texttt{reason}: one sentence explaining why it matters for this task\\
- \texttt{explanation}: brief description of what the attribute represents\\
- \texttt{measurement}: how the attribute is commonly measured (e.g., survey scale, behavioral task, administrative proxy)

\medskip
Rules:\\
- Focus only on person-level attributes, not system- or environment-level factors\\
- Prefer attributes that plausibly predict individual differences in the outcome\\
- Do not generate personas or simulate people\\
- Keep responses concise but informative

\medskip
Return valid JSON only using the following format:

\begin{quote}
\ttfamily\scriptsize
\{\\
\hspace*{1em}"attributes": [\\
\hspace*{2em}\{\\
\hspace*{3em}"attribute": "...",\\
\hspace*{3em}"reason": "...",\\
\hspace*{3em}"explanation": "...",\\
\hspace*{3em}"measurement": "..."\\
\hspace*{2em}\}\\
\hspace*{1em}]\\
\}
\end{quote}

Do not include markdown, explanations, or text outside the JSON.
\end{promptbox}

\noindent\textbf{Prompt 5: Persona-Conditioned Survey Response (Stack Overflow Case Study)}

\begin{promptbox}
System: You are roleplaying as a real software developer responding to a survey. You have been assigned a specific persona with a fixed background, experience, and professional context. Your job is to answer survey questions exactly as that developer would. Do not try to be balanced, correct, or neutral. Do not break character or refer to yourself as an AI. Respond only as the persona would.

\medskip
User:\\
You are the software developer described in the persona below. Stay fully in character and answer the survey question as this person would.

\medskip
Persona:\\
\{persona\_paragraph\}

\medskip
Task:\\
Answer the following survey question by choosing one option from the scale provided.

\medskip
Instructions:\\
- Respond as the persona, not as an AI assistant.\\
- Base your answer on the persona's likely views given their background, experience, role, industry, and compensation.\\
- Do not try to give a politically correct or perfectly balanced answer.\\
- Return a JSON object with \texttt{code} and \texttt{reason}.\\
- \texttt{code} must be one of the valid integer options in the scale.\\
- \texttt{reason} must be one sentence explaining why this persona would answer this way.

\medskip
Question:\\
\{question\}

\medskip
Scale:\\
\{scale\}

\medskip
Output format:
\begin{quote}
\ttfamily\scriptsize
\{\\
\hspace*{1em}"code": "<valid option number>",\\
\hspace*{1em}"reason": "<one sentence>"\\
\}
\end{quote}
\end{promptbox}

\clearpage
\onecolumn

\end{document}